# ESG Sentiment Analysis: comparing human and language model performance including GPT

Karim Derrick

University of Manchester

## Abstract

In this paper we explore the challenges of measuring sentiment in relation to Environmental, Social and Governance (ESG) social media. ESG has grown in importance in recent years with a surge in interest from the financial sector and the performance of many businesses has become based in part on their ESG related reputations. The use of sentiment analysis to measure ESG related reputation has developed and with it interest in the use of machines to do so.

The era of digital media has created an explosion of new media sources, driven by the growth of social media platforms. This growing data environment has become an excellent source for behavioural insight studies across many disciplines that includes politics, healthcare and market research.

Our study seeks to compare human performance with the cutting edge in machine performance in the measurement of ESG related sentiment. To this end researchers classify the sentiment of 150 tweets and a reliability measure is made. A gold standard data set is then established based on the consensus of 3 researchers and this data set is then used to measure the performance of different machine approaches: one based on the VADER dictionary approach to sentiment classification and then multiple language model approaches, including Llama2, T5, Mistral, Mixtral, FINBERT, GPT3.5 and GPT4.

## 1 Introduction

In recent years there has been a growth in interest in how corporates perform in respect of Environmental, Social and Governance issues inline with growing demand from both investors and the public for ESG investment funds (Schmidt 2019). At the same time media, in particular social media, has exploded and several studies have emerged that demonstrate a relationship between corporate ESG activities and the cost of corporate capital. The relationship between media coverage and corporate information transmission (Tetlock 2014) has long been established. More recent studies though have shown that corporate ESG activities matter in a number of ways, that positive

ESG activities affects media favourability (Tetlock 2005, Cahan, Chen et al. 2015) and, moreover, that ESG sentiment analysis can be used as a predictor for financial market sentiment. Positive ESG news builds investor trust and enhances reputation.

In this paper, we will consider how human assessment of ESG news sentiment impacts our ability to predict the impact of ESG issues on corporate stock value and corporate risk. We look to examine how effective humans are in the simple task of assessing the sentiment polarity of a sample of ESG related social media and we then examine the effectiveness of machine approaches to assessing the same social media. This leads us to two research questions for which this paper gives consideration:

**RQ1**: How consistent is human ESG sentiment analysis?

**RQ2**: Can the use of large language models produce better performance?

## 1.1 What is sentiment analysis?

Sentiment analysis is also known as opinion analysis, a study of opinions, attitudes and emotions towards objects or entities which might include products, organisations, stories, politics, or individuals. The analysis of text has a long history. Loughran and McDonald (2016) trace it back to the 1300 when friars first analysed biblical texts to index common phrases. In the world wars rhetorical choices were interpreted as diplomatic trends.

The term sentiment analysis first appeared in Nasukawa and Yi (2003) who developed a system to perform semantic analysis and recognised that the task requires "high intelligence and deep understanding of the textual context, drawing on common sense and domain knowledge as well as linguistics".

Examples of papers that have found it to be important include those looking at cyber security (Karimi, Vandam et al.) , analysis of political communications (Haselmayer and Jenny 2017, Song, Tolochko et al. 2020) and media impact on political preferences (Kleinnijenhuis, van Hoof et al. 2019).

The main objective in sentiment analysis is the identification of whether a text is positive or negative towards a subject. Measuring sentiment though is not straightforward with the expression of sentiment subjective and often ambiguous (Liu 2022).

## 1.2   Computing sentiment.

The most widely used methods of computing sentiment until only recently attempted to compute meaning from text relying on the assumption that the order and thus context of words is unimportant. These methods, typically known as "bag-of-words", reduce a document to a term matrix made up from rows of words and columns of word counts.

Sentiment dictionaries are a variation on the "bag-of-words" concept and instead consist of collections of words that look to determine a particular sentiment. The task is simple, make a long list of positive and negative words and then count how many words of each category occur. This has the advantage of ensuring researcher subjectivity and scales well with the size of the text as well as the number of samples (Loughran and McDonald 2016).

Dictionary based approaches generally do not perform well with many studies demonstrating an inter-coder reliability Krippendorf alpha of less than 0.4 (Van Atteveldt, Van Der Velden et al. 2021). Some dictionaries perform well when tailored for specific tasks in respect of their precision scores, but suffer from false negatives or type 2 errors impacting recall scores (Damstra and Boukes 2021). Applying a dictionary to investigate a specific research question can also lead to widely divergent conclusions as multiple studies have shown(Young and Soroka 2012).

Latent Semantic Analysis is a development of the simpler "bag-of_words" approach that uses singular value decomposition to reduce the term-document matrix dimensionality to isolate latent topics. In effect it is factor analysis for text.

Word classification using supervised machine learning is another method used with the most popular the Naıve Bayes method. These methods either use researcher given rules/filters or otherwise some other factor, such as the reaction to the text, as the dependent variable, thus removing the researcher's subjectivity. Numerous studies in the financial sector have used this approach to good effect as detailed by the Loughran and McDonald (2016) review, including Buehlmaier and Whited (2018) who investigated the impact of financial constraint on stock returns, and also Huang, Zang et al. (2014) who analysed 363,952 financial reports to demonstrate that they have predictive value for future earnings growth across the following 5 years. A Financial Phrasebank available at [financial phrasebank · Datasets at Hugging Face](financial phrasebank · Datasets at Hugging Face) was established by Malo, Sinha et al. (2014) and which has been used in multiple supervised machine learning studies using neural architectures that includes Maia, Freitas et al. (2018).

## 1.3 Language Models

Language models have been developed to predict the next word in a given text, or the next word given a text as context; as such they are a significant development in the natural language processing field. The basic models are trained on very large corpuses of text and are can then be fine-tuned on domain or task specific texts which can include semantic analysis. Fine-tuning language models allows the model weights to be adjusted for down-stream natural language tasks. Early attempts at this approach using BERT, the original Transformer model, include Araci (2019) which achieved state of the art classifications on the Financial Phrasebank from Malo, Sinha et al. (2014). More recent Large Language Models include the Generative Pretrained Transformers (GPT) which have been trained on 45TB of text and feature an incredible 175 billion parameters and have demonstrated incredible abilities out of the box. The first study of the ability of GPT in respect of sentiment analysis is Kheiri and Karimi (2023).

## 1.4 Estbablishing a Gold Standard

To evaluate machine-based approaches in sentiment analysis and to draw meaningful comparison against human performance, some form of objective measurement is required. In the literature, this is usually based on a gold standard with human judgement as its foundation. Ultimately, the assumption is that humans will make the "most correct and valid classifications of text" (Song, Tolochko et al. 2020). However, numerous studies referenced in this paper demonstrate that human judgement intercoder reliability is itself often poor, recognising that as with many areas of human judgement, insufficient attention is given to the task of determining what is acceptable when it comes to human classification and coding and most of the studies that this paper reference lack methodological detail of the validation methods that have been employed in order to establish their gold standard. Song, Tolochko et al. (2020) in their paper looked specifically at this issue and went on to model and thus demonstrate the implications of the variability in underlying gold standards. As machine models increase in efficacy, this methodological problem is being amplified and in turn bringing into question the use of human judgement as the gold standard. As this paper shows, with the performance of language models now able to produce output without configuration and training to a very high level, the use of human judgement as the gold standard may now be reaching a limit in utility.

## 1.5 Intercoder Reliability

All qualitative researchers using hand-coded data—that is, data in which items are labelled with categories, whether to support an empirical claim or to develop and test a computational model—should show that such data are reliable. If we are to use human judged ESG sentiment, we must ensure the reliability of our data. In practice, particular measures of reliability have been more popular in some analytic traditions over others. Content analysis, similar to sentiment analysis, in particular has developed a strong tradition for intercoder reliability measurement (O'Connor and Joffe 2020). According to (Neuendorf 2002), "Without the establishment of reliability, content analysis measures are useless". Kolbe and Burnett (1991) suggests that the quality of content research is a function of inter-coder reliability.

Reliability "is thus a prerequisite for demonstrating the validity of the coding scheme"; if it is to be demonstrated that the coding scheme genuinely captures the phenomenon being studied, reliability must be demonstrated. Qualitative research places value on the interpretation of data but does so with the expectation that the interpretation will be shared with others.

Intercoder reliability is often conflated with interrater reliability. Inter coder reliability refers to data rated on an ordinal or interval scale, whereas inter rater reliability is appropriate for data categorised on a nominal scale; the presence or absence of an emotion for example as opposed to the degree of an emotion. Most qualitative analysis will focus on intercoder reliability over interrater(O'Connor and Joffe 2020). In this study we are interested in sentiment analysis, and we are interested in whether sentiment is positive, negative or neutral, an ordinal classification and thus intercoder reliability,

Efforts to put semantics and discourse research on the same empirical footing as other areas of computational linguistics (CL) began in earnest in the 1990s and drove concern over the subjectivity of the human judgements required to create coded resources (Artstein and Poesio 2008). It was this concern that in turn drove the development of the original Cohen's κ (Cohen 1960) and the coefficients that followed. Prior to this development, analysis was largely limited to percentage agreement statistics.

The simplest measure of agreement between two coders is the percentage of agreement or observed agreement, defined for example by (Scott 1955)) as "the percentage of judgments on which the two analysts agree when coding the same data independently." The issues with this approach are multiple: firstly, some agreement will always be due to chance, secondly given multiple coding schemes examining the same phenomenon, the one with the fewest categories will likely result in the higher percentage agreement by virtue of chance. Percentage agreement does not consider the distribution of items among categories, categories that are common can expect more agreement. Articles surveyed between 1984 and 1987 found that 65% of computational linguistics articles used percentage agreement despite the serious flaws, and another study between 1981 and 1990 showed 78% using percentage agreement (Lombard, Snyder-Duch et al. 2002).

The evolution of the coefficients that are now regularly used for measuring inter-coder reliability, was originally driven by the need to account for chance and the distribution of items that had now been handled by percentage agreement. The three best known and most widely used coefficients, (BENNETT, ALPERT et al. 1954), Scott's Pi (Scott 1955), Cohen's Kappa (Cohen 1960) all have the same formula at their foundation:

$$S, \pi, K = \frac{A_0 - A_e}{1 - A_e}$$

- Ao is observed agreement.
- Ae is expected agreement.

All three coefficients assume independence of the two coders with the following variations:

- The S coefficient (BENNETT, ALPERT et al. 1954) assumes that if coders were operating by chance alone, we would get a uniform distribution. This is problematic in many respects. For example the value of the coefficient can be artificially increased simply by adding categories which the coders would never use (Artstein and Poesio 2008)
- The π coefficient (Scott 1955) assumes that if the coders were operating by chance alone, we would get the same distribution for each coder.
- The κ (Cohen 1960) assumes that If coders were operating by chance alone, we would get a separate distribution for each coder.

In corpus annotation practice, measuring reliability with only two coders is seldom considered enough, except for small-scale studies. Sometimes researchers run reliability studies with more than two coders, measure agreement separately for each pair of coders, and report the average. A generalization of Scott's π that copes with multiple coders is proposed in (Fleiss 1971)and a generalization of Cohen's κ is given in (Davies and Fleiss 1982). A serious limitation of both π and κ is that all disagreements are treated equally. But especially for semantic and pragmatic features, disagreements are not alike. Even for the relatively simple case of dialogue act tagging, a disagreement between an accept and a reject interpretation of a text is clearly more serious than a disagreement between an info-request and a check.

Di Eugenio and Glass (2004) noted the issue of the behaviour of agreement coefficients when the annotation data is severely skewed. The first issue, which Di Eugenio and Glass call the bias problem, shows that π and κ yield quite different numerical values when the annotators' marginal distributions are widely divergent; the second issue that they note is the prevalence problem, where it is difficult to get high agreement values when most of the items fall under one category.

The debates around the use of intercoder reliability indices drove Lombard, Snyder-Duch et al. (2004) to demand a standardisation of indices in content analysis. According to Lombard, Pasadeos, Huhman et al. (1995) found that in analyses of news-media messages between 1988 and 1993, 51% did not address reliability and 80% made no mention of coder training (Lombard, Snyder-Duch et al. 2002) – reliability is often seemingly an after thought despite it's importance in establishing the validity of a study.

This lead in turn to Lombard's recommendation that the Kirppendorf α is adopted (Krippendorff 1980, 2004a), which is a coefficient defined in a general way that is appropriate for use with multiple coders, different magnitudes of disagreement, and missing values, and is based on assumptions similar to those of π; and weighted kappa κw (Cohen 1968), a generalization of κ. For this study and for the reasons given above, we will use Krippendorff's alpha.

## 2 Method

For this study, in the first instance, 500 tweets were randomly collected from Twitter based only on whether or not those tweets contained a small number of preselected ESG related key words. Most tweets on the internet do no feature ESG related content, in order to test the effectiveness of human and machine approaches they will need to demonstrate an ability to discern subtle differences in ESG content and ESG intention and thus the initial selection was deliberately chosen to enable the ability for the approaches to discriminate to be fully tested. The tweets that were collected featured the selected words but were not necessarily the subject of any of them.

Having collected the tweets an initial exercise was conducted whereby four researchers were asked to classify each of 150 tweets based entirely on whether those tweets exhibited ESG sentiment, informed by Krippendorff's (Krippendorff 2019) recommendation that at least 143 documents are used for determining intercoder reliability in the context of media classification. The tweets were classified with -1 as negative, 1 as positive and zero as neutral. No other instruction was given and there was no opportunity for collaboration.

In the next phase a gold standard classification was established for the same 150 tweets in order to compare the different manual and automatic classification methods, similar to the comparative approach adopted by Van Atteveldt, Van Der Velden et al. (2021). Three experts were designated to establish the gold standard by review of the150 ESG related tweets. All disagreements were then discussed between the experts and resolved. As part of the proves, a simple coding scheme was created based on a mixed deductive and inductive approach. The scheme was established theoretically and was then refined by review of examples (Burla, Knierim et al. 2008).  The following coding rules were thus established between the experts, enabling agreement in respect of ESG specific sentiment:

- You are a media analyst. For the following tweet, would it reflect positively or negatively in respect of reputation, particularly in respect of ESG.
- A tweet is neutral if purely informational and just stating a fact and is not making a judgement, promoting or taking a stance.
- Tweets that directly or indirectly attack the ESG agenda are negative.

With the "gold standard" established, 4 machine based approached were then tested against the gold standard and a confusion matrix established for each method. The following methods were selected:

- Vader

  The only dictionary based approach selected, the Valence Aware Dictionary for Sentiment Reasoning is a python module that provides sentiment scores based on the words used. VADER was used in one of the earliest paper to explore the relationship between ESG news sentiment and stock price in Schmidt (2019).

- FinBERT

  FinBERT is a pre-trained NLP model to analyse sentiment of financial text built by fine-tuning the BERT language model using a large financial corpus and thereby fine-tuning it for financial sentiment classification. The [Financial PhraseBank](#) by Malo et al. (2014) is used for fine-tuning. The original BERT model was trained on 100million parameters in 2019..

- T5

    Developed by Google and was released in 2020, one year after the orginal BERT transformer and is a text-to-text transformer and has 220million parameters.

- GPT 3.5 Turbo

    The Generative Pre-trained Transformer 3.5 (GPT-3.5) large language model is a sub class of GPT-3 Models created by [OpenAI](OpenAI) in 2022 and rumoured to be trained on 175 billion parameters.

- GPT4

    At the time of writing GPT 4 is another multimodal large language model created by [OpenAI](OpenAI), and is the fourth and latest in its series of GPT foundation models and is rumoured to be trained on 1.76 trillion parameters.

- Llama 2 Chat

    This model was developed in a partnership between Microsoft and Meta, trained on 40% more data than Llama1. The 70B parameter version model was used in the test.

- Mistral 7B

    Mistral is a 7.3B parameter open-source model and at the time of writing has claimed improved performance over GPT 3.5 and Llama 2. It it said to use a similar archiecture to Llama but with changes to the attention mechanism.

- Mixtral 8x7B

    Mixtral uses a sparse mixture of experts architecture that provides the model with a total of 46.7B usable parameters.

## 3  Results

In the initial exercise 150 tweets were labelled by 4 researchers with no initial discussion, were then collated and an initial Krippendorf Alpha established. This initial study produced an alpha of only 0.3644 across the 900 pairs captured highlighting the poor level of reliability typically seen in labelling exercise where no coding standard is estbalished, highlighting the problem of using naïve human judges for tasks like this; out of the box, humans are poor sentiment labellers. Sentiment labelling is difficult enough, labelling with respect to the ESG agenda amplifies the level of error. Other studies utilising naive coders gave similarly poor results (Artstein and Poesio 2008).

The coding scheme that was then established removed a lot of the error seen in the initial study by bringing clarity to what is mean by ESG sentiment, particularly in respect of neutral informational statements. Iterations and discussion led the research team to conclude that for sentiment to be positive or negative, a stance, a judgement or a promotion had to be made in respect of either Environment, Social or Governance issues. Tweets that mention ESG activities are neutral if no stance is taken. In the following tweet reference is made to the need for companies to go beyond carbon offsetting in order to claim carbon neutrality, but no judgement is made, no stance is given; information is simply presented:

"There is a misconception within the financial sector that, through carbon offsetting alone, a company can make claims to carbon Neutrality or net-zero, Coad says."

Similarly occasional use of sarcasm can cause sentiment polarity to swing, in the following tweet the ESG agenda is attacked by sarcastic reference to a flight that Greta Thunberg had made, even though no reference to the flight is made:

"@GretaThunberg Curious, have you done the math yet on your carbon footprint going to Germany for this photo op?"

With the "gold standard" established, four machine methods were then evaluated with the gold standard used as the test data against the model's prediction. The following confusion matrices were produced.

|  | Accuracy | Recall | F1-score |
|---|---|---|---|
| GPT 4 | 0.95 | 0.95 | 0.95 |
| GPT 3.5 | 0.73 | 0.73 | 0.73 |
| Mixtral | 0.68 | 0.68 | 0.66 |
| LLAMA2 | 0.54 | 0.58 | 0.54 |
| T5 | 0.66 | 0.6 | 0.51 |
| FinBert | 0.48 | 0.48 | 0.48 |
| Mistral | 0.48 | 0.52 | 0.45 |
| Vader | 0.48 | 0.44 | 0.37 |
| BERT-FS | 0.43 | 0.38 | 0.33 |

Krippendorf alpha was calculated for 2 separate runs of GPT4 to check for consistency given the probabilistic foundation for the model's operation. An alpha of 0.9595 was achieved.

## 4 Discussion

As expected, the dictionary-based approach to sentiment analysis was less effective on nearly every measure. Just because media utterance features ESG related language does not mean that the polarity of the language used is a good predictor of the polarity of the overall statement. The dictionary approach cannot effectively discriminate and is close to chance agreement in respect of performance.

The large language model-based approaches all fared better. FinBERT performed least well. Good performance in identification of positive statements balanced by an inability to identify negative or neutral statements despite being trained specifically on a large financial statements corpus. This is a lot worse than the sentiment performance performed on financial statements without the ESG content observed by Araci (2019) which at the time exceeded the state of the art by 15%. Other studies (Li, Chan et al.) have suggested that FinBert can exceed OpenAi's models on financial tasks but this is when compared to performance specifically on the Financial Phrasebank (Malo, Sinha et al. 2014) for which FinBert has specifically been trained on

GPT3.5 was noticeable better but not stellar. Sarcasm was readily identified as noted by Kheiri and Karimi (2023). The ability to discern statements that are only informational is significantly improved over FinBERT but is not perfect. An informational statement noting an indirect impact of climate change without adopting a stance is thus incorrectly classified: "As climate change alters temperatures and weather patterns around the world, the risk of vector-borne diseases like dengue fever and Zika virus will increase."

GPT4 enjoys all the improvements of GPT3.5 but is better and more accurate on every count. Where GPT3.5 could not discern a subtle double negative, GPT 4 gets it right: "Twitter says it will no longer amplify accounts run by states 'engaged in armed interstate conflict' [https://t.co/1hLyQyXFDo](https://t.co/1hLyQyXFDo)".

Tweets that only indirectly take a stance by asking a question still indirectly imply a negative position. A subtlety identified by GPT4: "Are #CEOs and corporate executives still greenwashing or rainbow-washing their #ESG / #SDG goals? Some thoughts on this from my closing remarks at #GEPInnovate2022. [https://t.co/hgieVqoyQ4](https://t.co/hgieVqoyQ4)`". The big difference in performance of GPT4 and GPT3.5 when compared to FinBert on a sentiment classification task which is different but related to the Financial Phrasebank task for which FinBert was trained demonstrated how much more generalised the GPT family of models are.

Before Large Language Models and with human judgment as the "gold standard" against which everything else is measured, algorithms were deemed to be performing about as well as a random human judge but only when trained for a specific domain and usually at great effort; in other words, their performance was flawed.

With the step change in performance that these latest models achieve, does the use of large language models now that mean it is possible to exceed average human performance? Given the 95% level of accuracy of out of the box performance of GPT4 it seems likely that human performance will be exceeded without effort unless substantial effort is made to choose and train human judges. It is only with sunstantial effort and much discussion that the gold standard was established in this study. In comparison the language models are effortless. This has implications for the notion of human professionals: "Like a good measuring instrument, an expert judge must be both discriminating and consistent" (Weiss and Shanteau 2003). Many studies have demonstrated that professionals often lack consistency when it is needed most and for decades now it has been shown that algorithms can exceed professional performance. Meehl's seminal work demonstrated that statistical models were almost always more accurate than clinicians (Meehl 1954). Goldberg's work showed that a statistical model of an individual clinician was more accurate than the clinician modelled (Goldberg 1976). Until now, creating statistical models for human performance has been onerous requiring large datasets, much data wrangling, and a lot of compute power. The performance of out of the box GPT4 has exceeded humans in this study without any additional training at all, with minimal compute and with little setup overhead.

## 5 Contributions / implications

Most academic papers evaluating machines methods of sentiment analysis use human judgement as a gold standard; often the underlying methodology for determining the gold standard is largely hidden from view. Where more than one judge or expert is used, the requirement for meeting the gold standard is consensus. As we show here, naïve judges even when experts in their field will still be inconsistent without careful preparation. The latest language models achieve excellent correlation with our consensus-based gold standard whilst still ensuring a high level of consistency, the out of the box GPT4 by far exceeding the consistency of our out of the box researchers. It is less and less clear why humans would be used over a language model when this level of performance can be achieved and it also bring into question the use of human based gold standards for basic efficacy tests.

The Schmidt (2019) paper demonstrated that the use of the basic sentiment dictionary approach used by the VADER algorithm was still able to demonstrate "significant effects" of ESG-related news sentiment and stock market performance. In this paper we have demonstrated a significant improvement from GPT based models over VADER. The significant improvement in the accuracy of the sentiment measurement implies the potential for a significant improvement in the predictability of that stock movement and for which further study is required.

- The author of this report is an employee of Kennedys Law LLP which is developing products related to the research in this paper.